\documentclass{article}

\usepackage{PRIMEarxiv}

\usepackage[utf8]{inputenc}
\usepackage[T1]{fontenc}
\usepackage{hyperref}
\hypersetup{hidelinks}
\usepackage{url}
\usepackage{amsmath,amssymb,amsfonts}
\usepackage{graphicx}
\graphicspath{{./}}
\usepackage{textcomp}
\usepackage{xcolor}
\usepackage{booktabs}
\usepackage{multirow}
\usepackage{array}
\usepackage{placeins}
\usepackage{microtype}
\usepackage{nicefrac}

\newcommand{\citegap}{\nobreak\hspace{\fontdimen2\font}}

\title{MambaDSF: Multi-Scale State-Space Model with Dilated Feature Fusion for Sonar Small Target Detection}

\author{
  Hui Lin$^{1,*}$, Jiayi Li$^{1,*}$, Jing Wang$^{2}$, and Shenghui Rong$^{1,\dagger}$ \\
  $^{1}$School of Information Science and Engineering, Ocean University of China, Qingdao 266100, China \\
  $^{2}$School of Information and Communication Technology, Griffith University, Nathan, QLD 4111, Australia \\
  \texttt{\{harrylin929, leanolee58\}@gmail.com, wangjingname@gmail.com, rsh@ouc.edu.cn} \\
  {\small $^{*}$Equal contribution (co-first authors). $^{\dagger}$Corresponding author.}
}

\begin{document}
\maketitle
\vspace{-0.5cm}

\let\thefootnote\relax\footnotetext{Accepted for publication in \textit{IEEE Geoscience and Remote Sensing Letters}. \copyright~2026 IEEE. Personal use of this material is permitted. Permission from IEEE must be obtained for all other uses, in any current or future media, including reprinting/republishing this material for advertising or promotional purposes, creating new collective works, for resale or redistribution to servers or lists, or reuse of any copyrighted component of this work in other works.}

\begin{abstract}
Sonar imaging is the primary modality for underwater target
detection, yet small targets remain difficult to detect due
to insufficient pixel coverage, low acoustic contrast, and
scale ambiguity across imaging ranges.
CNN-based detectors extract local features efficiently
but cannot suppress noise-induced false alarms without
global acoustic context.
Transformer-based methods capture long-range dependencies
at quadratic computational cost. Existing Mamba-based
vision models offer efficient linear-cost scanning but
lack multi-scale
semantic alignment across pyramid levels,
multi-receptive-field fusion, and small-target-aware
training supervision needed for reliable sonar detection.
This letter proposes \textbf{Mamba}-based
\textbf{D}ilated \textbf{S}cale \textbf{F}usion
(\textbf{MambaDSF}), a hybrid framework addressing these
limitations through three contributions: a
\textbf{Mamba}-\textbf{E}nhanced \textbf{F}eature
\textbf{P}yramid (\textbf{MambaEFP}) backbone that preserves
weak echo and boundary cues while constructing multi-scale
candidate features; a \textbf{D}ilated \textbf{F}usion
\textbf{Mamba} (\textbf{DF-Mamba}) encoder that uses
long-range cross-scale state-space model (SSM) context to regularize candidate
responses across pyramid levels; and
\textbf{S}ize-\textbf{A}daptive
\textbf{W}asserstein-\textbf{IoU} (SA-WIoU) and
\textbf{C}ross-\textbf{S}cale
\textbf{C}oherence (CSC) losses that stabilize small-target
training. MambaDSF achieves 91.5\%
mAP$_{50}$ on the UATD forward-looking sonar benchmark
with 28.7\,M parameters, surpassing all compared
detectors. Among Mamba-based underwater detectors,
the gain on the UATD small-target subset reaches +7.7\,pp
in mAP$_{50\text{-}95}$, and cross-domain evaluation on
FLS and MD-FLS further suggests its robustness.
The codes are publicly available at
\url{https://github.com/IDontKnowAAA/MambaDSF}.
\end{abstract}


\section{Introduction}
\label{sec:intro}

\begin{figure}[t]
\centering
\includegraphics[width=\textwidth]{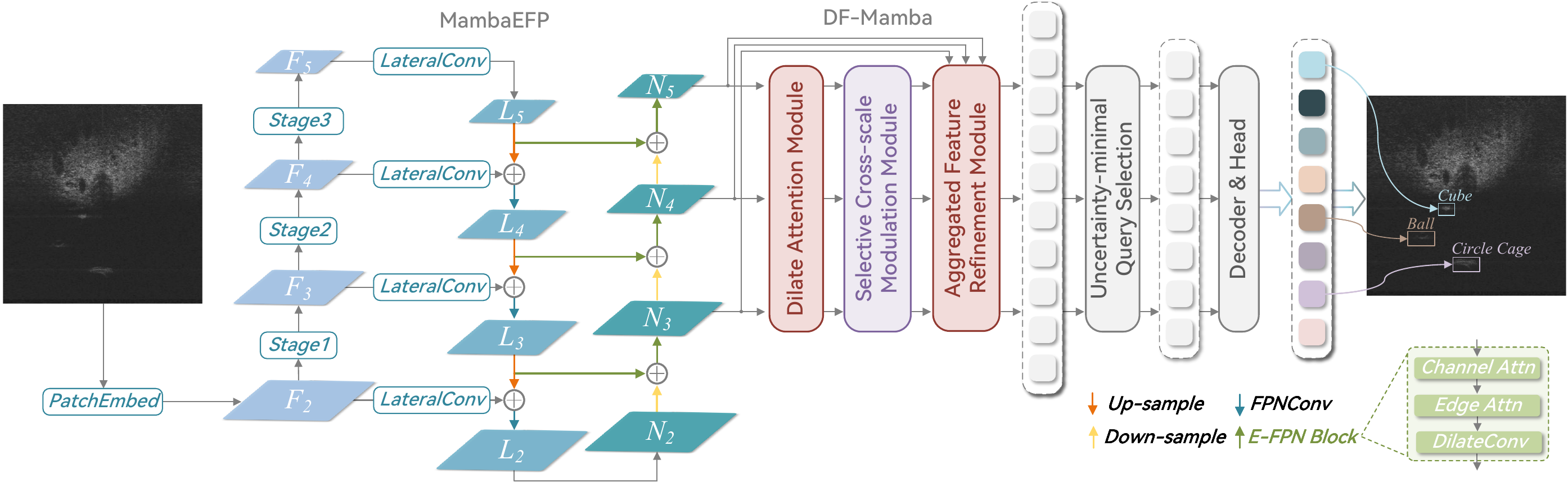}
\caption{Overall architecture of \textbf{Mamba}-based
\textbf{D}ilated \textbf{S}cale \textbf{F}usion
(\textbf{MambaDSF}). The \textbf{Mamba}-\textbf{E}nhanced
\textbf{F}eature \textbf{P}yramid (\textbf{MambaEFP}) backbone
extracts multi-scale features at three pyramid levels.
The \textbf{D}ilated \textbf{F}usion \textbf{Mamba}
(\textbf{DF-Mamba}) encoder refines each level via
multi-scale dilated attention and cross-scale
state-space model (SSM) fusion.
A query-based decoder yields detections
from 300 learnable queries.}
\label{fig:pipeline}
\end{figure}

Sonar imaging serves as the primary
modality for underwater target detection in marine survey,
port security, and autonomous underwater vehicle (AUV)
navigation\citegap\cite{zhou2022automatic}.
Forward-looking sonar (FLS) and side-scan sonar (SSS) are
widely deployed for real-time frontal imaging and
high-resolution seabed mapping,
respectively. Both modalities share
fundamental challenges for small targets.

Sonar targets face detection difficulties analogous to
optically small objects because the underwater acoustic
channel introduces propagation loss, multipath interference,
and scattering, producing low spatial
resolution, strong speckle noise, and blurred target
boundaries. Small sonar targets
consequently generate weak echoes readily masked by
background reverberation, and their edges tend to merge
with the surrounding seabed\citegap\cite{xie2022uatd}.

Three interconnected challenges arise from these
conditions. \textbf{Scarce discriminative features}:
targets may collapse to one or two feature-map cells
after successive
downsampling\citegap\cite{xie2022uatd}.
\textbf{Low target-to-clutter contrast}: echo intensities
from true targets are comparable to reverberation and
speckle, causing frequent false
alarms.
\textbf{Scale ambiguity}: apparent target size shifts
with imaging range and grazing
angle\citegap\cite{zhou2022automatic}, demanding feature
representations across multiple receptive-field scales.
Taken together, these conditions define the task of sonar 
small target detection as localizing and classifying every 
target in a forward-looking sonar image with a bounding 
box and label, emphasizing targets spanning only tens of pixels.

CNN-based detectors\citegap\cite{yolov8,ren2015faster} extract local features efficiently but
cannot model the global acoustic noise pattern needed to
separate true targets from pervasive clutter.
Transformer-based methods, including
RT-DETR\citegap\cite{zhao2024rtdetr},
and the sonar-oriented LS-DETR\citegap\cite{wang2025lsdetr},
achieve global context modeling via self-attention yet
incur quadratic complexity that limits real-time
deployment. SSMs, particularly Mamba\citegap\cite{gu2024mamba},
aggregate spatial context at \emph{linear} cost through
input-dependent selective gating, making them well suited
to clutter discrimination.
MambaVision\citegap\cite{hatamizadeh2024mambavision} extends
SSMs to hierarchical vision backbones, and
MiM-ISTD\citegap\cite{chen2024mimistd} demonstrates their
advantage for infrared small targets. However, 1-D
flattening of 2-D feature maps dilutes sparse target
tokens, and no existing Mamba-based detector enforces
cross-scale semantic consistency, provides
multi-receptive-field coverage, or includes
small-target-specific supervision for sonar.
Recent infrared small-target methods, including DFAwareNet
\citegap\cite{cheng2026dfaware}, WaveTD
\citegap\cite{ma2025wavetd}, and HFMNet
\citegap\cite{cheng2025hfmnet}, enhance weak targets through
dynamic frequency cues, wavelet-guided features, and lightweight
feature enhancement, respectively, but do not explicitly address
sonar-specific speckle, echo-shadow ambiguity, and
range-dependent scale variation. MFPD\citegap\cite{guo2026mfpd} 
uses Mamba-driven feature-pyramid decoding for underwater 
object detection, yet its generic fusion is insufficient 
for scarce-pixel sonar targets.

To overcome these limitations, we propose
\textbf{Mamba}-based \textbf{D}ilated
\textbf{S}cale \textbf{F}usion (\textbf{MambaDSF}), a hybrid framework with three
contributions. \textbf{(1)}~We design the \textbf{Mamba}-\textbf{E}nhanced \textbf{F}eature
\textbf{P}yramid (\textbf{MambaEFP}) backbone to adapt MambaVision to
sonar weak-target perception: the \textbf{Hybrid Block} preserves
two-dimensional local echo structure, while the enhanced feature
pyramid (E-FPN) preserves weak
contrast, edge, and localization cues before they are attenuated
during deeper encoding.
\textbf{(2)}~We introduce the \textbf{D}ilated \textbf{F}usion
\textbf{Mamba} (\textbf{DF-Mamba}) encoder, which goes beyond
ordinary pyramid resampling by using Multi-Scale Dilated Attention
(MSDA) for range-dependent receptive fields and Selective
Cross-scale Modulation (SCM) to condition Fusion State-Space Model
(FusSSM) on adjacent-scale
acoustic evidence, evaluating candidate reliability with long-range
cross-scale context. \textbf{(3)}~We devise \textbf{S}ize-\textbf{A}daptive
\textbf{W}asserstein-\textbf{IoU} (SA-WIoU) and
\textbf{C}ross-\textbf{S}cale \textbf{C}oherence (CSC) losses
to stabilize scarce-pixel box regression and cross-scale target
representation.

\section{Proposed Method}
\label{sec:method}

\subsection{Architecture Overview}
\label{subsec:overview}

Sonar small target detection localizes and classifies targets that 
span only tens of pixels in forward-looking sonar images, where 
scarce discriminative features, low target-to-clutter contrast, 
and scale ambiguity complicate detection. MambaDSF follows a three-stage 
pipeline in Fig.~\ref{fig:pipeline}. From a forward-looking sonar image
resized to $640\times640$, the \textbf{MambaEFP} backbone
extracts hierarchical features $\{F_i\}_{i=3}^{5}$ via
MambaVision, projects them into lateral features
$\{L_i\}_{i=3}^{5}$, and refines them through a Hybrid Block
and enhanced bidirectional pyramid to preserve weak echo,
boundary, and localization cues, yielding $\{N_i\}_{i=3}^{5}$ at strides
\{8,16,32\} with 256 channels each. The \textbf{DF-Mamba}
encoder takes $\{N_i\}_{i=3}^{5}$ and applies multi-scale dilated
attention and cross-scale SSM fusion at each level
to evaluate candidate reliability with long-range
cross-scale context and suppress clutter-induced responses, producing
$\{E_i\}_{i=3}^{5}$. A query-based decoder produces $300$ predictions, 
each comprising class scores over the target categories and a bounding box,
supervised by two task-specific losses: \textbf{SA-WIoU} counters
gradient vanishing for small targets, and \textbf{CSC} enforces
semantic consistency across levels.

\subsection{MambaEFP Backbone}
\label{subsec:backbone}

MambaEFP integrates
a MambaVision feature extractor, a Hybrid Block, and an
enhanced feature pyramid (E-FPN).

\textbf{MambaVision feature extractor.}
MambaVision\citegap\cite{hatamizadeh2024mambavision} is a
hierarchical network whose later stages use
Mamba-Transformer blocks with selective SSM,
where $\mathbf{h}_t=\bar{\mathbf{A}}\,\mathbf{h}_{t-1}
+\bar{\mathbf{B}}\,\mathbf{x}_t$
and $\mathbf{y}_t=\mathbf{C}\,\mathbf{h}_t
+\mathbf{D}\,\mathbf{x}_t$.
Here, $\bar{\mathbf{A}}\!=\!\exp(\Delta\mathbf{A})$,
$\bar{\mathbf{B}}\!=\!\Delta\mathbf{B}$, and
$\mathbf{B}$, $\mathbf{C}$, $\Delta$ are input-dependent.
The input-dependent $\Delta$ is central to sonar
processing: large values retain target token information in
the hidden state while small values suppress noise tokens, providing scene-wide clutter attenuation at
linear cost. The hierarchical features are extracted
across the backbone stages.

\begin{figure}[t]
\centering
\includegraphics[width=0.66\columnwidth]{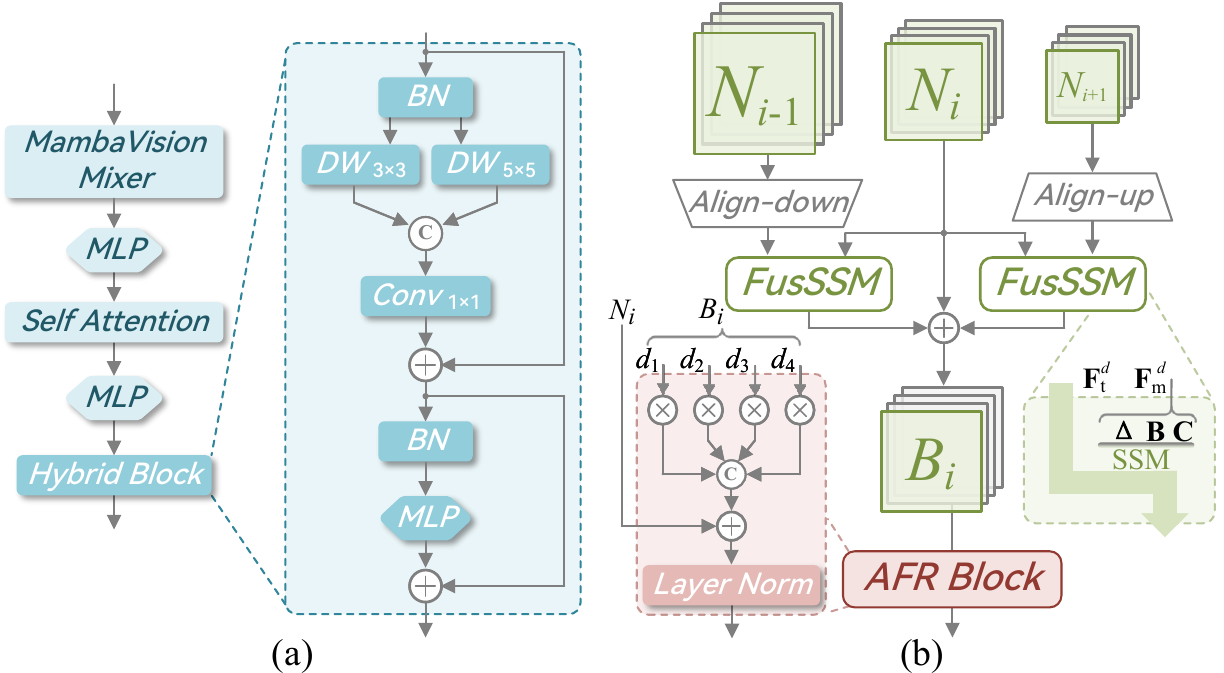}
\caption{Detailed module architectures. (a)~Hybrid Block:
parallel DW$_{3\times3}$ and DW$_{5\times5}$ branches
preserve local spatial detail alongside the SSM branch,
merged via learnable residual weights. (b)Selective
Cross-scale Modulation (SCM) and Aggregated Feature
Refinement (AFR) Blocks: Multi-Scale Dilated Attention
(MSDA) splits features into $n_d$ branches; Fusion
State-Space Model (FusSSM)
derives SSM parameters $(\Delta,\mathbf{B},\mathbf{C})$
from aligned auxiliary features and applies multiplicative
gating for cross-scale fusion.}
\label{fig:blocks}
\end{figure}

\textbf{Hybrid Block.}
Although MambaVision accumulates global context
efficiently, flattening 2-D feature maps into 1-D sequences
dilutes sparse target tokens within the dominant
background. The Hybrid Block, inserted at
Stage~3 (Fig.~\ref{fig:blocks}(a)), addresses this through
a dual-branch design: a \textit{local branch} with
parallel DW$_{3\times3}$ and DW$_{5\times5}$ depthwise
convolutions operates directly in the spatial domain,
preserving fine-grained echo boundaries without sequence
flattening; its output is combined with the SSM branch
through learnable residual weights, allowing the network
to adaptively balance local precision and global
awareness.

\textbf{Enhanced bidirectional feature pyramid network (E-FPN).}
Sonar targets are distinguished by intensity contrast and
boundary sharpness rather than semantic texture, demanding
explicit reinforcement of these cues before cross-scale
aggregation. E-FPN combines three lightweight 
operations per block: channel attention for echo-intensity contrast,
depthwise edge attention for high-frequency boundaries, and
ASPP-style dilation for echo envelopes of different extents.
Together, these operations increase sensitivity to weak target
candidates and preserve their localization cues before the subsequent
DF-Mamba encoder evaluates candidate reliability with cross-scale
context. PANet\citegap\cite{liu2018panet} adds a bottom-up path for
localization feedback. The outputs $\{N_i\}_{i=3}^{5}$, each
with 256 channels, are passed to the DF-Mamba encoder.

\subsection{Dilated Fusion Mamba Encoder (DF-Mamba)}
\label{subsec:encoder}

\begin{figure}[t]
\centering
\includegraphics[width=0.9\textwidth]{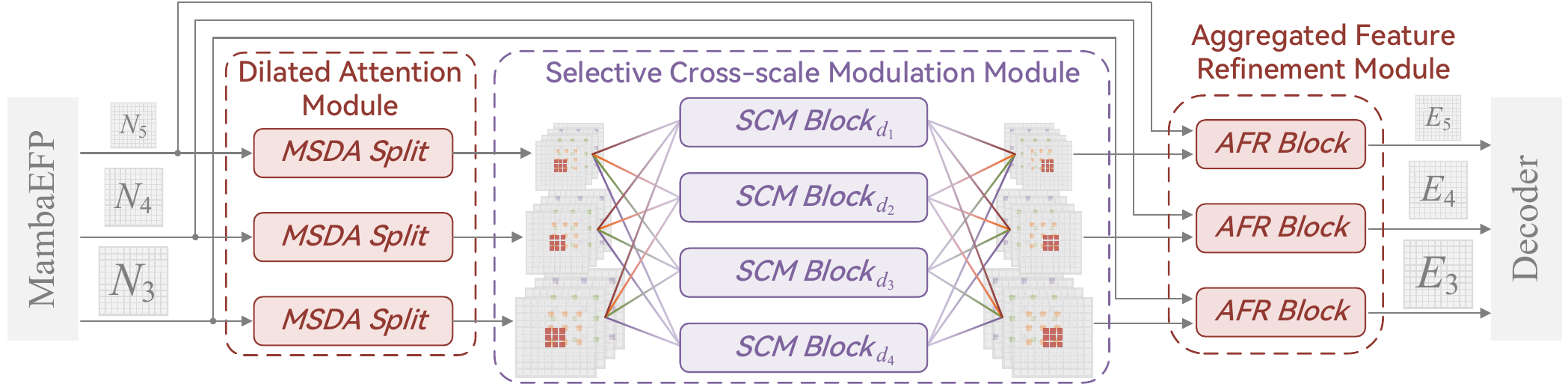}
\caption{\textbf{D}ilated \textbf{F}usion \textbf{Mamba}
(\textbf{DF-Mamba}) encoder architecture. $\{N_3,N_4,N_5\}$
from the \textbf{Mamba}-\textbf{E}nhanced \textbf{F}eature
\textbf{P}yramid (\textbf{MambaEFP}) are split into $n_d$
dilated attention branches by the \textbf{M}ulti-\textbf{S}cale
\textbf{D}ilated \textbf{A}ttention (\textbf{MSDA}) module.
Each branch undergoes independent cross-scale fusion in the
\textbf{S}elective \textbf{C}ross-scale \textbf{M}odulation
(\textbf{SCM}) Block via the \textbf{F}usion
\textbf{S}tate-\textbf{S}pace \textbf{M}odel
(\textbf{FusSSM}), whose gating parameters derive from
spatially aligned auxiliary features of adjacent pyramid levels.
The \textbf{A}ggregated \textbf{F}eature \textbf{R}efinement
(\textbf{AFR}) module aggregates all branches and produces
$\{E_3,E_4,E_5\}$ for the decoder.}
\label{fig:encoder}
\end{figure}

MambaEFP produces candidate-sensitive multi-scale representations,
yet the encoder further refines two key properties. \textit{Multi-receptive-field sensitivity} is
needed because the echo envelope of different sonar target
categories spans substantially different spatial extents.
\textit{Cross-scale semantic coherence with long-range dependency modeling} is required because
acoustic shadow length, range-dependent background clutter, and highlight intensity vary with imaging
conditions, causing the same target or target-like response to elicit
inconsistent responses across pyramid levels. DF-Mamba
resolves these requirements through a two-step process.

\textbf{Intra-scale enhancement via Multi-Scale Dilated
Attention (MSDA)}\citegap\cite{jiao2023dilateformer}.
At each level $N_i$, query, key, and value tensors are
split into $n_d$ groups of $C/n_d$ channels, each
processed at a distinct dilation rate $d$:
\begin{equation}
\text{DilAttn}_d\!=\!
\text{softmax}\!\!\left(\!
\frac{\mathbf{Q}^d\!\cdot\!
\text{Unfold}_d(\mathbf{K}^d)^\top}{\sqrt{C/n_d}}
\!\right)\!\text{Unfold}_d(\mathbf{V}^d),
\label{eq:dil_attn}
\end{equation}
where $\text{Unfold}_d$ extracts a $3\!\times\!3$
neighbourhood at pixel spacing $d$, giving branch $d$ an
effective receptive field of
$(2d\!+\!1)\!\times\!(2d\!+\!1)$. Smaller $d$ preserves
fine boundary detail while larger $d$ captures the broader
echo envelope, enabling joint encoding of local target cues
and surrounding context. Branch $d$ outputs the
target feature \mbox{$\mathbf{F}_{\mathrm{t}}^d$}.

\textbf{Cross-scale fusion via Selective Cross-scale
Modulation (SCM) and Aggregated Feature Refinement
(AFR).}
Each branch $d$ undergoes independent cross-scale fusion
in the SCM Block. Auxiliary features from adjacent pyramid
levels are spatially aligned and projected to form
modulator $\mathbf{F}_{\mathrm{m}}^d$, which is flattened into four
directional sequences and processed by
Fusion State-Space Model (FusSSM)\citegap\cite{gao2025msfmamba},
whose parameters derive from the modulator. Thus,
$\mathbf{F}_{\mathrm{m}}^d$ conditions the scan of
$\mathbf{F}_{\mathrm{t}}^d$, so the gate verifies whether a candidate
has adjacent-scale support and intra-scale long-range context. 
SCM suppresses isolated high-response clutter lacking
cross-scale support, and AFR aggregates the modulated branches for
the decoder.

A learnable projection $\mathbf{W}_p$ maps the modulator to
$\Delta_\text{fus},\,\mathbf{B}_\text{fus},\,\mathbf{C}_\text{fus}
=\text{Split}\!\big(\mathbf{W}_p\,\mathbf{F}_{\mathrm{m}}^d\big)$.
$\mathbf{F}_{\mathrm{t}}^d$ is then scanned by
the selective SSM recurrence with these parameters, and the four
directional outputs form a sigmoid gate:
\begin{equation}
\mathbf{g}\!=\!\sigma\!\Big(
\textstyle\sum_{k=1}^{4}\text{SSM}_k\!\big(
\mathbf{F}_{\mathrm{t}}^d;\,\bar{\mathbf{A}},\,
\bar{\mathbf{B}}_\text{fus}^{(k)},\,
\mathbf{C}_\text{fus}^{(k)}\big)\Big),
\label{eq:fus_gate}
\end{equation}
The modulated output
$B_i^d=(1\!-\!\alpha)\,\mathbf{F}_{\mathrm{t}}^d
+\alpha\,(\mathbf{F}_{\mathrm{t}}^d\odot\mathbf{g})$,
where $\alpha$ is learnable, amplifies positions with
consistent multi-level support because $\mathbf{g}$ is derived from
the modulator rather than the target branch. AFR then concatenates
all branches with a residual skip:
$E_i\!=\!\text{LN}(\text{Concat}(B_i^{1..n_d})
\!+\!N_i)$.

\begin{table}[t]
\centering
\caption{Comparison across four evaluation 
benchmarks. All models were pretrained on UATD and fine-tuned 
on each target dataset under identical settings. Best results 
in \textbf{bold}, second-best \underline{underlined}.}
\label{tab:comparison}
\renewcommand{\arraystretch}{1.1}
\setlength{\tabcolsep}{1.6pt}
\small
\resizebox{\textwidth}{!}{%
\begin{tabular}{llccc cc cc cc cc}
\toprule
& & & &
& \multicolumn{2}{c}{\textbf{UATD Test}}
& \multicolumn{2}{c}{\textbf{UATD Subset}}
& \multicolumn{2}{c}{\textbf{FLS Test}}
& \multicolumn{2}{c}{\textbf{MD-FLS Test}} \\
\cmidrule(lr){6-7} \cmidrule(lr){8-9} 
\cmidrule(lr){10-11} \cmidrule(lr){12-13}
\textbf{Type} & \textbf{Method}
& \textbf{Params} & \textbf{GFLOPs} 
& \textbf{FPS}
& \textbf{mAP$_{50}$} & \textbf{mAP$_{50\text{-}95}$}
& \textbf{mAP$_{50}$} & \textbf{mAP$_{50\text{-}95}$}
& \textbf{mAP$_{50}$} & \textbf{mAP$_{50\text{-}95}$}
& \textbf{mAP$_{50}$} & \textbf{mAP$_{50\text{-}95}$} \\
\midrule
\multirow{5}{*}{CNN}
& Faster R-CNN\citegap\cite{ren2015faster}
  & 41.4\,M & 65.4
  & 96.6
  & 86.8 & 39.7
  & 49.5 & 20.7
  & 69.0 & 37.4
  & 89.1 & 57.5 \\
& YOLOv3\citegap\cite{redmon2018yolov3}
  & 61.6\,M & 283.0
  & 116.2
  & 85.6 & 39.6
  & 52.9 & 26.8
  & 71.4 & 38.4
  & 89.7 & 63.1 \\
& YOLOv5-L\citegap\cite{jocher2020yolov5}
  & 53.2\,M & 135.3
  & 123.3
  & 87.1 & 41.9
  & 50.4 & 26.9
  & 75.1 & 43.0
  & 88.9 & 62.6 \\
& YOLOv8-L\citegap\cite{yolov8}
  & 43.6\,M & 165.4
  & 137.9
  & 87.2 & 44.6
  & 56.5 & 27.5
  & 78.7 & 47.5
  & 89.5 & 64.4 \\
& YOLOv10-L\citegap\cite{wang2024yolov10}
  & 25.8\,M & 127.3
  & 90.8
  & 86.8 & 41.4
  & 53.7 & 24.7
  & 73.1 & 47.1
  & 85.5 & 62.4 \\
\midrule
\multirow{4}{*}{Trans.}
& Def.\ DETR\citegap\cite{zhu2021deformabledetr}
  & 41.6\,M & 160.7
  & 31.1
  & 88.7 & 42.7
  & 40.5 & 19.4
  & 81.9 & 49.0
  & \textbf{92.3} & 65.9 \\
& DINO-DETR\citegap\cite{zhang2022dino}
  & 47.6\,M & 119.0
  & 41.1
  & 82.8 & 40.1
  & 56.1 & 25.1
  & 77.4 & 51.5
  & 88.9 & 66.2 \\
& RT-DETR R50\citegap\cite{zhao2024rtdetr}
  & 42.9\,M & 135.8
  & 67.6
  & \underline{89.2} & 43.5
  & 56.3 & \underline{28.8}
  & 77.2 & 52.3
  & 90.8 & 65.3 \\
& LS-DETR\citegap\cite{wang2025lsdetr}
  & 13.7\,M & 45.3
  & 37.4
  & 86.1 & 42.5
  & \underline{56.6} & 26.1
  & 80.6 & 50.8
  & 89.7 & 66.4 \\
\midrule
 \multirow{2}{*}{SSM}
& MFPD\citegap\cite{guo2026mfpd}
& 33.0\,M & 102.3
  & 46.2
  & 89.0 & \underline{44.8}
  & 53.4 & 22.6
  & \underline{82.5} & \underline{54.8}
  & 90.0 & \underline{68.1} \\
& \textbf{MambaDSF}
  & 28.7\,M & 141.5
  & 47.5
  & \textbf{91.5} & \textbf{46.8}
  & \textbf{58.8} & \textbf{30.3}
  & \textbf{83.3} & \textbf{56.7}
  & \underline{91.0} & \textbf{69.2} \\
\bottomrule
\end{tabular}%
}
\end{table}

\subsection{Decoder and Loss Function}
\label{subsec:decoder}

Small sonar targets often produce zero IoU between
predicted and ground-truth boxes, causing standard
regression gradients to vanish. Meanwhile, independent
per-level supervision allows contradictory representations
of the same target across encoder scales. We introduce two
task-specific losses to address these issues. Detections
are generated by the adopted RT-DETR
decoder\citegap\cite{zhao2024rtdetr} operating on
$\{E_3,E_4,E_5\}$. The total loss is
$\mathcal{L} = \mathcal{L}_{\text{focal}}
+ \lambda_s \mathcal{L}_{\text{SA-WIoU}}
+ \ell_1
+ \lambda_c \cdot \mathcal{L}_{\text{CSC}}$.

\textbf{Size-Adaptive Wasserstein-IoU (SA-WIoU).}
Each bounding box $\mathbf{b}\!=\!(c_x,c_y,w,h)$ is
modelled as a 2-D Gaussian with mean
$(c_x,c_y)^\top$ and covariance
$\mathrm{diag}(w^2/4,h^2/4)$. The squared Wasserstein
distance between predicted and ground-truth Gaussians
admits the closed form\citegap\cite{wang2021nwd}:
\begin{equation}
\mathcal{W}_2^2 = \|\boldsymbol{\mu}_p-\boldsymbol{\mu}_g\|^2
+ \tfrac{1}{4}\bigl[(w_p\!-\!w_g)^2+(h_p\!-\!h_g)^2\bigr],
\label{eq:w2}
\end{equation}
yielding the Normalized Wasserstein Distance
$\mathcal{L}_\text{NWD}\!=\!
1\!-\!\exp(-\mathcal{W}_2/\tau_w)$.
SA-WIoU blends this with CIoU\citegap\cite{zheng2020ciou} via an
area-adaptive weight:
\begin{equation}
\mathcal{L}_{\text{SA-WIoU}}
= \omega\,\mathcal{L}_{\text{NWD}}
+ (1\!-\!\omega)\,\mathcal{L}_{\text{CIoU}},\;
\omega = \exp\!\Bigl(-\frac{a_g}{\tau_s}\Bigr),
\label{eq:sawiou}
\end{equation}
where $a_g$ is the normalized ground-truth box area,
so that $\omega$ automatically favours the Wasserstein term
for small targets and CIoU for large ones.

\textbf{Cross-Scale Coherence (CSC).}
Feature vectors sampled at each ground-truth box centre
across $\{E_3,E_4,E_5\}$ are aligned in \emph{direction}
via pairwise cosine similarity, unifying categorical
identity without constraining per-level magnitudes:
\begin{equation}
\mathcal{L}_{\text{CSC}}
= 1 - \frac{1}{|\mathcal{P}||\mathcal{G}|}
\sum_{g\in\mathcal{G}}\sum_{(i,j)\in\mathcal{P}}
\frac{\mathbf{e}_i^g \cdot \mathbf{e}_j^g}{\|\mathbf{e}_i^g\|\,\|\mathbf{e}_j^g\|},
\label{eq:csc}
\end{equation}
where $\mathcal{P}\!=\!\{(3,4),(4,5),(3,5)\}$ and
$\mathcal{G}$ is the set of ground-truth boxes in the
current batch. SA-WIoU and CSC jointly stabilize training
by preserving regression gradients and resolving cross-scale
semantic conflicts, respectively.

\section{Experiments}
\label{sec:exp}

\subsection{Experimental Setup}

We evaluated MambaDSF on three sonar
benchmarks. The UATD dataset\citegap\cite{xie2022uatd} contains 9,200
multibeam forward-looking sonar images across 10 categories
(7:1:2 train/val/test split) and serves as the primary
training source. A small-target subset is further
constructed by selecting annotations whose bounding-box
area falls below the 25th percentile of the UATD test-set
distribution ($\approx$50$\times$50\,pixels).
FLS\citegap\cite{zhu2021fls} and
MD-FLS\citegap\cite{valdenegro2025mdfls} are two
additional forward-looking sonar datasets used exclusively
for cross-domain evaluation: models pretrained on UATD were
fine-tuned on each target set with the backbone frozen
for 50 epochs.
All experiments used an NVIDIA RTX\,4090 GPU,
$640\!\times\!640$ input resolution, and the AdamW
optimizer. We report mAP$_{50}$ and mAP$_{50\text{-}95}$
following the COCO protocol.

\subsection{Comparison with State-of-the-Art}
\label{subsec:sota}

\begin{figure}[t]
\centering
\includegraphics[width=\textwidth]{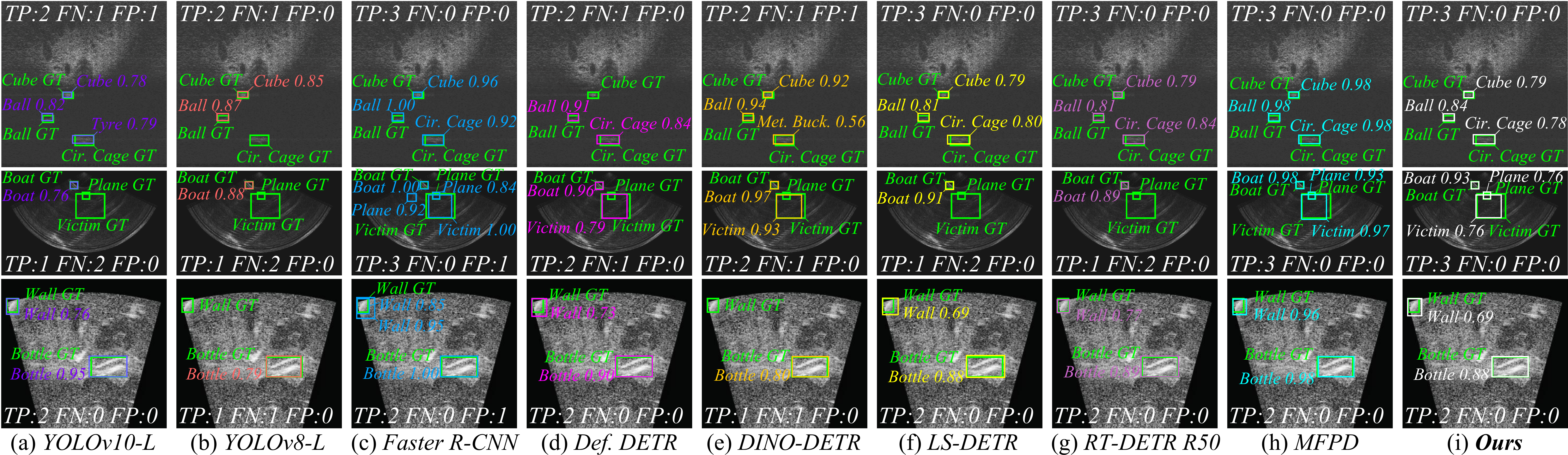}
\caption{Qualitative comparison across three sonar
datasets (one row per dataset: UATD, FLS, MD-FLS) and
nine compared methods.
Ground-truth boxes are shown in green; TP/FN/FP counts are
annotated in each sub-image.}
\label{fig:detection_grid}
\end{figure}

Table~\ref{tab:comparison} summarizes the four
evaluation settings. On the full UATD test set,
MambaDSF achieves the highest mAP$_{50}$ of 91.5\% and
mAP$_{50\text{-}95}$ of 46.8\%, surpassing
the compared CNN-, Transformer-, and Mamba-family detectors.
Compared with MFPD, MambaDSF improves
mAP$_{50\text{-}95}$ by 2.0\,pp on full UATD. 
MambaEFP preserves weak echo and boundary cues for
low-contrast targets, while DF-Mamba combines multi-receptive-field
attention with cross-scale SSM fusion to regularize candidate responses
and suppress clutter-induced false positives.
On the small-target subset, MambaDSF retains the highest
accuracy across both metrics, reaching 58.8\% mAP$_{50}$ and
30.3\% mAP$_{50\text{-}95}$, which is 7.7\,pp higher than
MFPD in mAP$_{50\text{-}95}$. This supports the claim that global context
modeling, multi-scale alignment, and gradient-preserving regression
jointly benefit scarce-pixel target detection.
Cross-domain results on FLS and MD-FLS further suggest its robustness:
MambaEFP preserves weak echo and boundary cues under
domain-varying imaging conditions, DF-Mamba regularizes candidate
responses with multi-scale and long-range context, and
SA-WIoU sharpens localization, yielding the highest mAP$_{50}$ on
FLS and the highest mAP$_{50\text{-}95}$ on both cross-domain sets.
In terms of efficiency, MambaDSF achieves 47.5\,FPS with 28.7\,M
parameters, more compact than 
eight of the ten baselines; its
141.5\,GFLOPs and lower throughput relative to CNN-based methods
reflect the added cost of SSM recurrence and multi-scale fusion, yet
this computational trade-off yields the highest accuracy across all
evaluation settings.

\subsection{Ablation Study}
\label{subsec:ablation}

\begin{table}[t]
\centering
\caption{Component ablation on the UATD test set.
\textbf{A}: MambaEFP backbone vs.\ vanilla MambaVision;
\textbf{B}: full DF-Mamba encoder vs.\ MSDA only;
\textbf{C}: proposed losses (SA-WIoU\,\&\,CSC)
vs.\ focal\,+\,GIoU\,+\,$\ell_1$.}
\label{tab:ablation}
\renewcommand{\arraystretch}{1.1}
\setlength{\tabcolsep}{1.8pt}

\begin{tabular}{>{\centering\arraybackslash}p{0.60cm}
                >{\centering\arraybackslash}p{0.60cm}
                >{\centering\arraybackslash}p{0.60cm}|
                >{\centering\arraybackslash}p{0.85cm}
                >{\centering\arraybackslash}p{0.85cm}
                >{\centering\arraybackslash}p{1.15cm}
                >{\centering\arraybackslash}p{1.15cm}}
\toprule
\textbf{A} & \textbf{B} & \textbf{C}
& \textbf{P} & \textbf{R}
& \textbf{\mbox{\scriptsize mAP$_{50}$}} & \textbf{\mbox{\scriptsize mAP$_{50\text{-}95}$}} \\
\midrule
 & & & 91.4 & 94.4 & 88.2 & 42.9 \\
 & & \checkmark & 91.0 & 94.6 & 88.6 & 44.0 \\
 & \checkmark & & 93.4 & 95.0 & 89.7 & 45.2 \\
\checkmark & & & 93.7 & 95.0 & 89.0 & 45.2 \\
 & \checkmark & \checkmark & 92.5 & 95.0 & 89.9 & 45.8 \\
\checkmark & & \checkmark & 93.1 & 95.2 & 89.2 & 45.3 \\
\checkmark & \checkmark & & 94.2 & 95.9 & 90.1 & 46.3 \\
\checkmark & \checkmark & \checkmark
  & 93.7 & 95.5 & \textbf{91.5} & \textbf{46.8} \\
\bottomrule
\end{tabular}
\end{table}

Table~\ref{tab:ablation} presents a $2^3$ full factorial
ablation of the three proposed components. Each single-component
addition improves over the baseline. Component~B yields the largest
single-factor mAP$_{50}$ gain, confirming that multi-receptive-field
coverage is the dominant factor for handling diverse sonar echo sizes.
Component~A delivers a comparable mAP$_{50\text{-}95}$ improvement,
reflecting the benefit of 
weak-cue preservation and global context for clutter suppression that vanilla MambaVision
alone cannot provide due to 1-D serialization dilution. Component~C primarily boosts
mAP$_{50\text{-}95}$, consistent with SA-WIoU's gradient advantage
for tightly localizing small targets. Two-factor combinations
consistently exceed their single-factor counterparts, confirming
that the three components address distinct aspects and combine
synergistically. The full model achieves 91.5\% mAP$_{50}$ and
46.8\% mAP$_{50\text{-}95}$.

\subsection{Qualitative Results}
\label{subsec:qualitative}

Fig.~\ref{fig:detection_grid} presents representative detection results
across all three datasets. CNN-based methods frequently miss
low-contrast targets or generate false positives where target echo
intensity approaches the background level, a pattern consistent across
all three datasets and attributable to their inability to model global
acoustic clutter.
Transformer-based methods perform more reliably on
MD-FLS but struggle on the multi-scale FLS scene, where
targets span a wider range of apparent sizes.
MFPD provides a strong Mamba-family comparison and also performs well in the displayed cases. MambaDSF nevertheless maintains clean detections with no additional false positives in these examples, consistent with its aggregate advantage reported in Table~\ref{tab:comparison}.


\section{Conclusion}
\label{sec:conclusion}

This letter proposes MambaDSF, a hybrid SSM-Transformer
framework for small target detection in sonar imagery,
built around three coordinated contributions: the
MambaEFP backbone for joint local and global feature
extraction, the DF-Mamba encoder for multi-scale semantic
alignment, and task-specific SA-WIoU and CSC losses for
stable small-target supervision. On UATD, MambaDSF achieves 91.5\% mAP$_{50}$ and
46.8\% mAP$_{50\text{-}95}$
with 28.7\,M parameters, and reaches
58.8\% mAP$_{50}$ and 30.3\% mAP$_{50\text{-}95}$ on the
UATD small-target subset. Cross-domain evaluation on FLS and
MD-FLS indicates that the learned representations generalize
across sonar imaging geometries. Ablation validates that the
three components address distinct aspects of the detection
problem and combine synergistically. 
Future work will further evaluate deployment on AUV
platforms and extend the framework to broader underwater perception tasks.

\section*{Acknowledgments}
This work was supported in part by the National Natural
Science Foundation of China under Grant 62001443, and in part by
the Natural Science Foundation of Shandong Province under Grant
ZR2020QE294.

\bibliographystyle{unsrt}
\bibliography{references}

\end{document}